\documentclass[runningheads]{llncs}

\usepackage{accv}

\usepackage{accvabbrv}

\usepackage{graphicx}
\usepackage{booktabs}
\usepackage{multirow}
\usepackage[accsupp]{axessibility}  

\usepackage[pagebackref,breaklinks,colorlinks,citecolor=accvblue]{hyperref}
\usepackage{hyperref}

\usepackage{orcidlink}

\begin{document}

\title{Video Domain Incremental Learning for Human Action Recognition in Home Environments}

\titlerunning{Abbreviated paper title}

\author{
    Yuanda Hu\inst{1} \and
    Xing Liu\inst{1} \and
    Meiying Li\inst{1} \and
    Yate Ge\inst{1} \and
    Xiaohua Sun\inst{2} \and \\
    Weiwei Guo\inst{1}\thanks{Corresponding author: Weiwei Guo, \email{weiweiguo@tongji.edu.cn}}
}

\authorrunning{Yuanda Hu et al.}

\institute{
    College of Design and Innovation, Tongji University, China\\
    \email{\{ydhu,lotsoliu,mzlzyCA,geyate,weiweiguo\}@tongji.edu.cn}
    \and
    SUSTech School of Design, Southern University of Science and Technology, China\\
    \email{sunxh@sustech.edu.cn}
}

\maketitle

\begin{abstract}
It is significantly challenging to recognize daily human actions in domestic settings due to the diversity and dynamic changes in unconstrained home environments. It spurs the need to continually adapt to various users and scenes. Fine-tuning current video understanding models on newly encountered domains often leads to catastrophic forgetting, where the models lose their ability to perform well on previously learned scenarios. To address this issue, we formalize the problem of Video Domain Incremental Learning (VDIL), which enables models to learn continually from different domains while maintaining a fixed set of action classes. Existing continual learning research primarily focuses on class-incremental learning, while the domain incremental learning has been largely overlooked in video understanding. In this work, we introduce a novel benchmark of domain incremental human action recognition for dynamic residential environments. We design three domain split types (user, scene, hybrid) to systematically assess the challenges posed by domain shifts in real-world settings. Furthermore, we propose a baseline learning strategy based on replay and reservoir sampling techniques without domain labels to handle scenarios with limited memory and task agnosticism. Extensive experimental results demonstrate that our simple sampling and replay strategy outperforms most existing continual learning methods across the three proposed benchmarks.

\end{abstract}

\section{Introduction}
\begin{figure}[tb]
  \centering
  \begin{subfigure}[b]{\textwidth}
    \centering
    \includegraphics[width=0.6\textwidth]{img/intro1.pdf}
    \caption{}
    \label{fig:intro1}
  \end{subfigure}
  
  \begin{subfigure}[b]{\textwidth}
    \centering
    \includegraphics[width=0.7\textwidth]{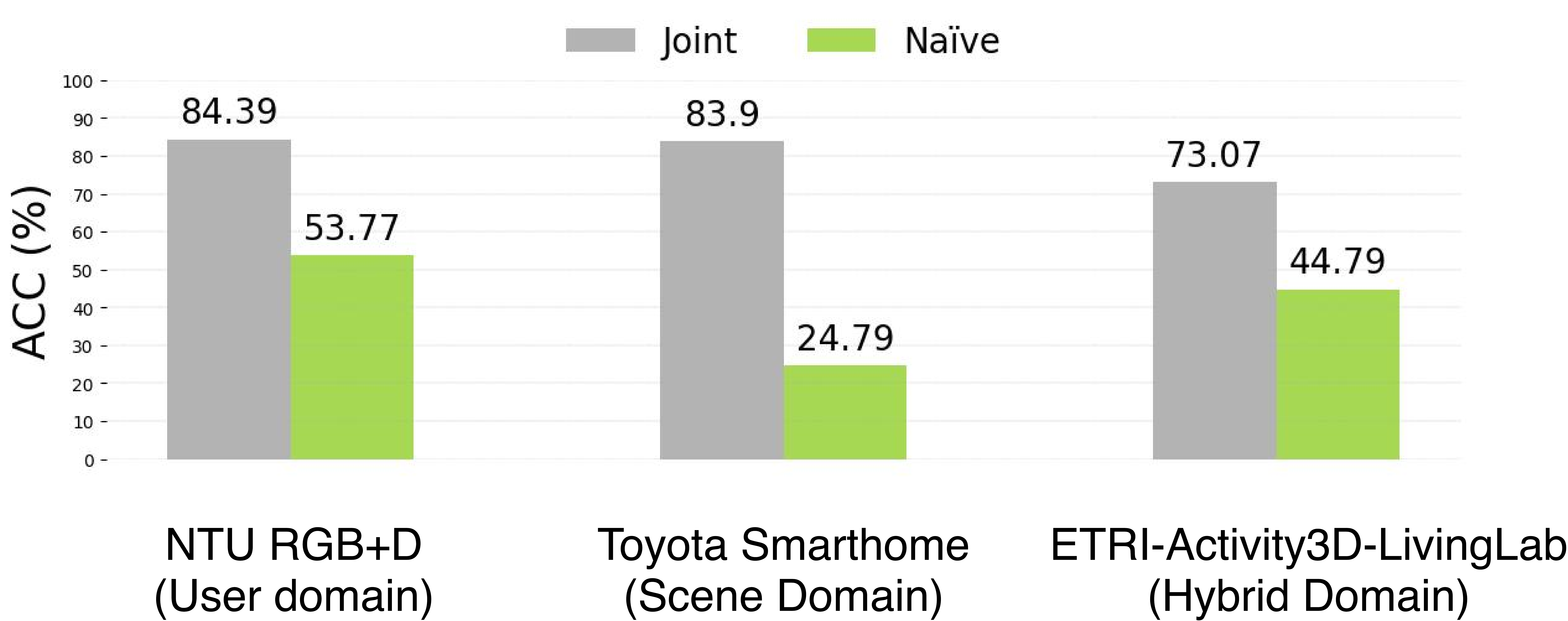}
    \caption{}
    \label{fig:intro2}
  \end{subfigure}
  
  \caption{(a) Comparison of Video Class Incremental Learning (VCIL) and Video Domain Incremental Learning (VDIL). VCIL learns new classes incrementally with a fixed data distribution, while VDIL learns from evolving data distributions within a constant class set. (b) The bar plots show the presence of catastrophic forgetting in VDIL paradigm, indicated by the performance degradation on previous tasks when learning unseen domains sequentially.}
  \label{fig:intro}
\end{figure}

Daily human action recognition in home environments has a wide range of applications, including service robots and smart living homes\cite{DBLP:journals/ijcv/KongF22}. Recent advancements in deep neural networks have propelled video understanding tasks to unprecedented levels of performance, largely due to the utilization of extensive and diverse datasets\cite{DBLP:journals/corr/abs-1212-0402,DBLP:conf/iccv/GoyalKMMWKHFYMH17,DBLP:conf/cvpr/HeilbronEGN15}. However, these models often exhibit poor generalization when confronted with scenarios that deviate from their pre-training data distribution, necessitating ad-hoc retraining\cite{alssumJustGlimpseRethinking2023a,DBLP:conf/cvpr/YangHSS22}. Video human action recognition in home environments faces challenges due to diverse and dynamic changes in home environments, such as illumination changes, changing backgrounds, and users\cite{DBLP:journals/tnn/XuCMCXY24,DBLP:journals/corr/abs-2211-10412,DBLP:journals/corr/abs-2006-03876}. Furthermore, user privacy concerns, limited on-device storage capacity, and high data annotation costs render continuous data collection and model retraining impractical. Consequently, sequential re-training of large-scale models becomes imperative\cite{DBLP:journals/inffus/LesortLSMFR20}. Video understanding models often face new action domains sequentially with limited access to previous data. In these scenarios, naive fine-tuning approaches result in catastrophic forgetting of previously learned tasks\cite{DBLP:journals/neco/FrenchC02}. This phenomenon manifests as a severe performance degradation on earlier tasks as the model adapts to new domains, compromising its ability to retain and apply prior knowledge. Furthermore, these models suffer from performance deterioration when encountering training data that deviates from the previously estimated distribution\cite{DBLP:conf/cvpr/RoyerL15}.

Continual learning has emerged as a promising approach to mitigate catastrophic forgetting, enabling models to adapt to novel patterns while maintaining performance on previously learned tasks \cite{delangeContinualLearningSurvey2021}. In video understanding, current research predominantly focuses on class-incremental learning, where new classes are progressively introduced \cite{masanaClassIncrementalLearningSurvey2022,delangeContinualLearningSurvey2021,masanaClassIncrementalLearningSurvey2022}. However, real-world scenarios present not only class increments but also continual domain shifts within existing classes. These domain shifts significantly alter the visual representations of actions and objects, necessitating adaptive models capable of handling evolving data distributions\cite{DBLP:journals/ml/Ben-DavidBCKPV10}. Our investigation reveals a critical phenomenon in home environments: user-specific behaviors and scene variations across different rooms and camera viewpoints significantly contribute to catastrophic forgetting, as illustrated in \cref{fig:intro2}. We frame this phenomenon within the continual learning paradigm as a Domain-Incremental Learning (DIL) problem, where the class set remains constant across tasks, but the distribution of these classes evolves \cite{vandevenThreeTypesIncremental2022}, illustrated in \cref{fig:intro1}. However, current research lacks comprehensive benchmarks to evaluate the effectiveness of models in addressing these domain-specific variations, particularly in dynamic home environments.

\begin{figure}[tb]
  \centering
  \includegraphics[width=1.0\linewidth]{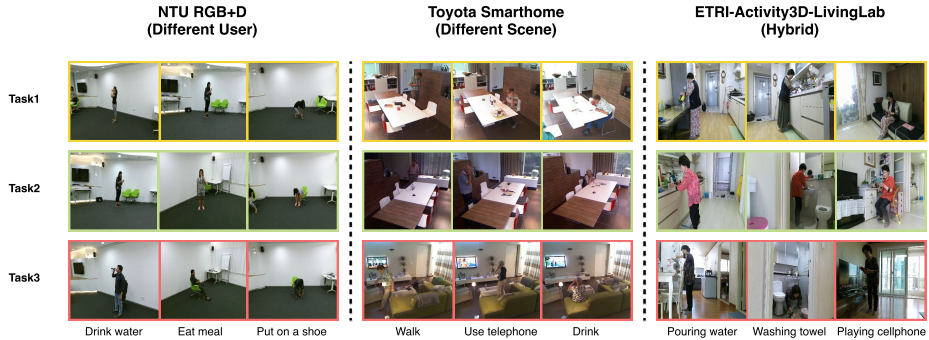}
  \caption{Illustration of the Video Domain Incremental Learning (VDIL) benchmarks, where models learn from evolving data distributions while the action classes remain fixed. We split three datasets (NTU RGB+D, Toyota Smarthome, and ETRI-Activity3D-LivingLab) into different domain types: user, scene, and hybrid (combining user and scene), to create benchmarks that evaluate models' ability to incrementally adapt to domain shifts commonly encountered in home scenarios.}
  \label{fig:benchmark}
\end{figure}

To address these critical issues, our work focuses on the DIL paradigm within continual learning, which aims to effectively handle evolving domains and temporal shifts in datasets. We introduce a novel benchmark for Video Domain Incremental Learning (VDIL) in home environments. Our benchmark incorporates three distinct domain split types. The user-based split captures variations in action performance across individuals, accounting for personalized styles and habits. The scene-based split focuses on environmental changes, including rooms and camera viewpoints, which can alter the visual representation of actions. The hybrid split combines both user and scene variations, simulating complex real-world scenarios where multiple factors contribute to domain shift, as shown in \cref{fig:benchmark}. These carefully designed split types enable a comprehensive examination of domain shift challenges in authentic home settings.

While sharing commonalities with class incremental learning, domain incremental learning for video understanding in home environments presents a set of unique obstacles. These include the rigidity-plasticity dilemma \cite{DBLP:conf/cvpr/KimH23}, continual domain drift due to evolving visual representations\cite{DBLP:conf/mm/0001JMLC22}, memory constraints limiting on-device storage\cite{DBLP:journals/inffus/LesortLSMFR20}, and the absence of clear task boundaries in real-world scenarios\cite{buzzegaDarkExperienceGeneral2020}. To tackle these challenges, we propose a baseline learning strategy that leverages replay and reservoir sampling techniques. Our results demonstrate a competitive performance compared with non-incremental and traditional incremental learning models across three benchmarks.

Our main contributions are summarized as follows:
\begin{itemize}
\item We formally define the problem setting of Video Domain-Incremental Learning (VDIL), addressing challenges posed by evolving domains in daily human action recognition within dynamic home environments.
\item We introduce a novel benchmark for domain incremental video human action recognition, incorporating user-based, scene-based, and hybrid domain splits. This benchmark enables quantitative analysis of catastrophic forgetting under various domain shift scenarios.
\item We propose a baseline method that leveraging replay and reservoir sampling techniques, designed to handle limited memory and task-agnostic scenarios in VDIL, demonstrating competitive performance. 
\end{itemize}

To the best of our knowledge, this is the first work to explore domain incremental learning in video understanding. Our approach demonstrates significant potential for robust home action recognition systems, enabling adaptation to diverse domestic environments and individual user behaviors while maintaining consistent performance.

\section{Related Work}
\subsection{Continual Learning in Video Understanding}
In the field of video continual learning, current leading methods merge regularization and exemplar-based approaches to better process temporal information. TCD \cite{parkClassIncrementalLearningAction2021} assesses the importance of the time channel to enhance weighted distillation, ensuring the retention of essential temporal information in all tasks. \cite{zhaoWhenVideoClassification2021a} explores spatio-temporal knowledge to impose stricter constraints on knowledge transfer, thereby increasing the model's resilience to forgetting. Subsequent studies \cite{alssumJustGlimpseRethinking2023a,villaVCLIMBNovelVideo2022a,peiLearningCondensedFrame2022} utilize knowledge distillation to ensure alignment between downsampled or condensed frames and the original video frames. Leveraging visual language pre-trained models has emerged as a powerful strategy in continual learning, providing a robust foundation for integrating new knowledge without substantial retraining.  PIVOT \cite{villaPIVOTPromptingVideo2023a}, minimizing the number of trainable parameters, incorporates insights from pre-trained models, while ST-Prompt \cite{peiSpacetimePromptingVideo2023} introduces task-specific and task-agnostic prompts, offering a rehearsal-free approach to the unique challenges of Video-Class Incremental Learning (VCIL). Although current studies primarily focus on class-incremental learning, our research uniquely addresses the unexplored area of domain incremental learning in video understanding, a significant gap for practical applications in real-world settings.

\subsection{Domain Incremental Learning}
Domain-incremental learning (DIL) is a subset of continual learning where the distribution of instances from fixed classes changes between domains \cite{vandevenThreeTypesIncremental2022}. Unlike task-incremental learning, which uses task indices during training and testing \cite{delangeContinualLearningSurvey2021}, and class-incremental learning, which integrates new classes over time \cite{masanaClassIncrementalLearningSurvey2022}, DIL focuses on adapting to shifts in input distribution while the set of classes remains unchanged \cite{wangSPromptsLearningPretrained2022,lamersClusteringbasedDomainIncrementalLearning2023,gargMultiDomainIncrementalLearning2022,jehanzebmirzaEfficientDomainIncrementalLearning2022,shiUnifiedApproachDomain2023,alssumJustGlimpseRethinking2023a}. The primary challenge is the dual objective of adapting to new instance distributions while retaining performance on previous distributions, thereby mitigating catastrophic forgetting. Research on domain-incremental learning (DIL) largely focuses on image classification.  Replay-based methods  mitigate catastrophic forgetting by using stored \cite{rebuffiICaRLIncrementalClassifier2017, chaudhryTinyEpisodicMemories2019, buzzegaDarkExperienceGeneral2020} or generated \cite{shinContinualLearningDeep2017} examples from previous tasks during training. In contrast, regularization-based methods apply constraints to network parameters to preserve previously learned information \cite{kirkpatrickOvercomingCatastrophicForgetting2017}, often leveraging techniques like knowledge distillation \cite{ rebuffiICaRLIncrementalClassifier2017, liLearningForgetting2018, wuLargeScaleIncremental2019}. Architectural strategies, on the other hand, redesign the network to be expandable, allowing it to adapt to new tasks while retaining knowledge from past tasks \cite{mallyaPackNetAddingMultiple2018, mallyaPiggybackAdaptingSingle2018}. This paper addresses the previously unexplored problem of domain-incremental learning for human action recognition, focusing on developing models that can adapt to ever-changing environmental conditions without prior knowledge of task identity during both training and testing phases.

\subsection{Unsupervised Domain Adaptation}
Loosely related to our work are Video Unsupervised Domain Adaptation (VUDA) methods, which aims to develop transferable video classification models for unlabeled target domains by minimizing discrepancies between source and target domains \cite{panAdversarialCrossDomainAction2020,yangInteractAlignLeveraging2022}, or by learning domain-invariant representations \cite{songSpatiotemporalContrastiveDomain2021,turrisidacostaDualHeadContrastiveDomain2022}.  However, VUDA primarily concentrates on enhancing performance in target domains. In contrast, our research is closely linked with Continuous Domain Adap tation (CDA), where a series of discrete adaptation steps are executed across multiple domains \cite{wangConfidenceAttentionGeneralization2023}. Unlike traditional domain adaptation that often relies on source domains to compensate for limited target data, our method assumes no such scarcity and focuses on consistent performance across all domains. Moreover, our framework is designed to tackle the challenge of adapting to an indefinite number of evolving, unseen target domains. 

\section{Video Domain Incremental Learning Benchmark}
\subsubsection{Notation \& Problem Definition}
Define \(\mathcal{D}_{\text{train}} = \{\mathcal{D}_k\}_{k=1}^T\) as the set of domain-incremental training tasks for video classification, where each task \(\mathcal{D}_k\) consists of samples \(\{(V_{k,j}, y_{k,j})\}_{j=1}^{|\mathcal{D}_k|}\), with each \(V_{k,j}\) representing the \(j\)-th video and \(y_{k,j}\) its corresponding label from a fixed set of classes \(\mathcal{C}\). Each task \(k\) encompasses all classes in \(\mathcal{C}\) and presents a unique data distribution specific to its domain.

Each task \(k\) corresponds to a distinct domain \(d_k\), with data distributions specific to \(d_k\), ensuring that \(d_k \cap d_{k'} = \emptyset\) for \(k \neq k'\). During the training phase for task \(k\), only the current data \(\mathcal{D}_k\) is accessible. Additionally, a limited memory buffer \(\mathcal{B}\) is permitted to store previous samples and their corresponding information, where the buffer size \(|\mathcal{B}|\) is significantly smaller than \(|\mathcal{D}_k|\) for all \(k \in \{1, 2, \ldots, T\}\).

\subsubsection{The Goal of DI Classification}
The overarching goal of video domain-incremental classification is to maximize the model's predictive accuracy across all encountered domains while minimizing the forgetting of previously learned information. Mathematically, this can be expressed as the following optimization problem:

\begin{equation}
\min_\theta \sum_{k=1}^T \mathcal{L}_k, \quad \text{where} \quad \mathcal{L}_k = \mathbb{E}_{(V_{k,j}, y_{k,j}) \sim \mathcal{D}_k, \mathcal{B}} \left[\ell(y_{k,j}, f_\theta(V_{k,j}))\right]
\end{equation}

Here, \(f_\theta(V_{k,j})\) represents the model with parameters \(\theta\) and \(\ell\) denotes the loss function. The parameters \(\theta\) are updated sequentially on each task to optimize the model \(f_\theta\). The objective is to find the set of parameters \(\theta\) that minimizes the loss across all tasks \(k \in \{1, \ldots, T\}\).

\subsubsection{Metrics of VDIL.} 
In the context of Video Domain Incremental Learning (VDIL), we employ two key metrics to evaluate model performance: Average Classification Accuracy (AA) and Backward Forgetting (BWF).

The AA measures the average accuracy over all tasks learned so far. It is computed as:

\begin{equation}
\text{AA} = \frac{1}{T} \sum_{k=1}^{T} a_{k,T}
\end{equation}

where \( T \) is the total number of tasks, \( a_{k,T} \) is the accuracy of the model on task \( k \) after training on task \( T \).

The BWF quantifies the amount of forgetting that occurs in the model. It is computed as the average decrease in performance on the previous tasks. The formula for BWF is:

\begin{equation}
\text{BWF} = \frac{1}{T-1} \sum_{k=1}^{T-1} \left( a_{k,k} - a_{k,T} \right)
\end{equation}

where \( T \) is the total number of tasks.\( a_{k,k} \) is the accuracy of the model on task \( k \) immediately after training on task \( k \). \( a_{k,T} \) is the accuracy of the model on task \( k \) after training on the final task \( T \).

\subsection{Benchmark Construction for VDIL}

Unlike traditional action recognition datasets that cover a wide range of scenarios, we focus specifically on home environments to construct our VDIL benchmark using three datasets: NTU RGB+D \cite{nturbgd}, Toyota Smarthome \cite{toyota}, and ETRI-Activity3D-LivingLab \cite{etri}. As shown in Table 1, we strategically divide these datasets to represent distinct domain shift scenarios common in domestic settings.

\textbf{NTU RGB+D} containing over 56,880 action samples captured from 40 subjects using RGB-D cameras \cite{nturbgd}. This dataset is characterized by its diverse collection of single-person and multi-person actions, recorded across a fixed set of scenes with varying camera distances. Despite the consistent laboratory setting, the dataset's strength lies in its subject diversity. We leverage this feature to construct our VDIL benchmark by defining domains based on different users, a method we term as \textit{Users Domain} division. Each domain encompasses a distinct set of users, allowing us to evaluate the model's capacity for continual learning across diverse individual action styles and patterns. 

\textbf{Toyota Smarthome} featuring 31 activities performed by 16 subjects in an authentic smart home environment \cite{toyota}. A key characteristic of this dataset is its comprehensive coverage of domestic scenes, captured by 7 strategically placed camera viewpoints throughout the house. This multi-view setup provides a rich representation of various home settings and environmental conditions. We exploit this feature to construct our VDIL benchmark by defining domains based on different rooms and camera perspectives, an approach we term as \textit{Scene Domain} division. Each domain represents a distinct combination of room layout and camera angle, enabling us to evaluate the model's ability to continually adapt to and learn from diverse spatial contexts.

\textbf{ETRI-Activity3D-LivingLab} captured in authentic home environments with diverse subjects and rooms \cite{etri}. Unlike datasets with fixed scenes, ETRI-Activity3D-LivingLab uniquely features multiple scenes for each user, closely mimicking real-world variability. We leverage this distinctive characteristic to construct our VDIL benchmark by defining domains based on both users and rooms, an approach we term as \textit{Hybrid Domain} division. Each domain encapsulates a combination of user-specific action patterns and diverse scene information, providing a more complex and realistic learning scenario. This hybrid approach allows us to evaluate the model's capacity to simultaneously adapt to varying user behaviors and environmental conditions.

\begin{figure}[tb]
  \centering
  \includegraphics[width=0.5\linewidth]{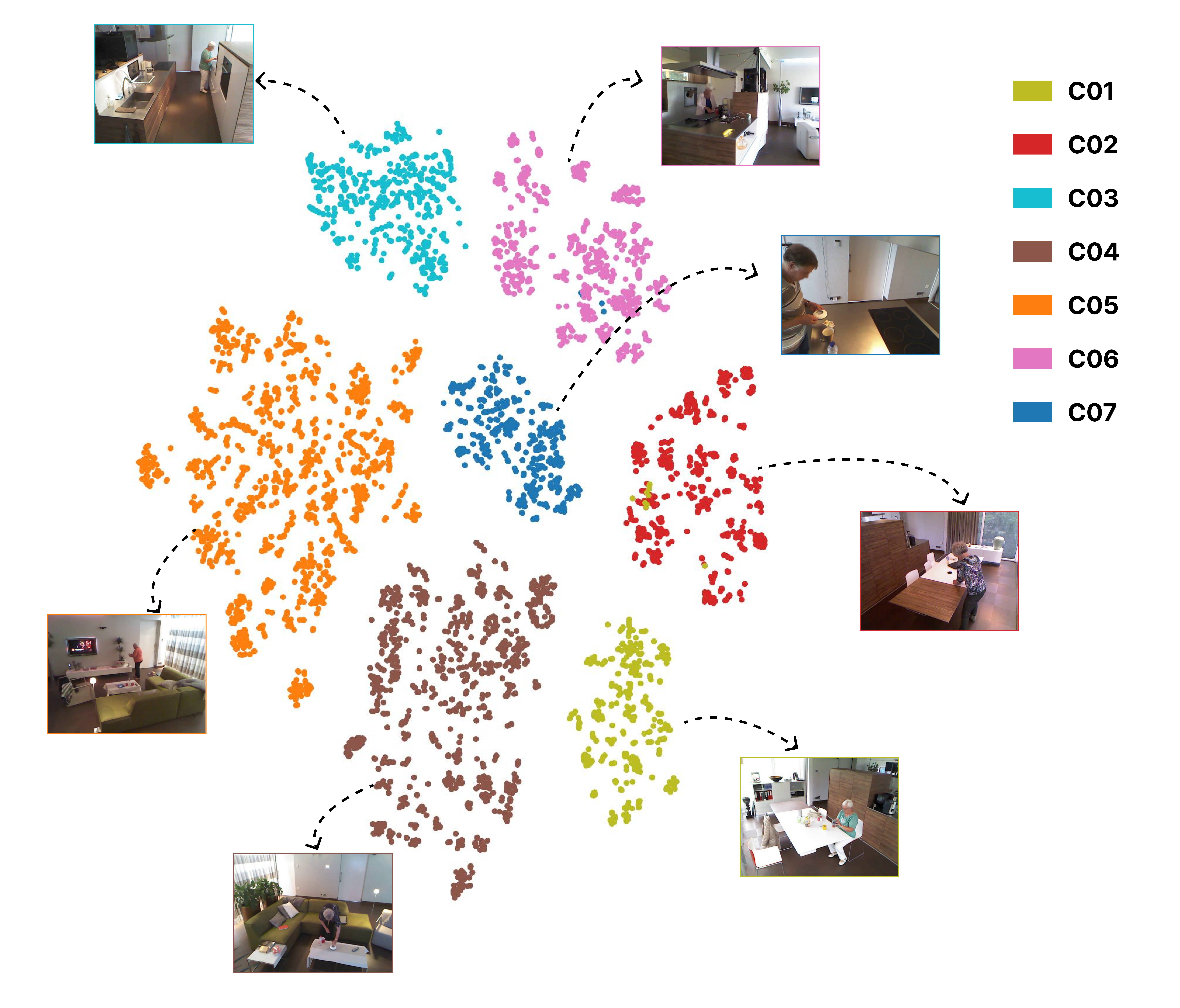}
  \caption{T-SNE visualization of feature distributions in the Toyota Smarthome dataset\cite{toyota}. Colors represent distinct \textit{Scene Domains}, each corresponding to different camera viewpoints. Contrary to expectations, samples cluster by domain rather than action class.
  }
  \label{fig:tsne}
\end{figure}

\begin{table}[tb]
  \caption{Video Domain Incremental Learning (VDIL) Benchmark Datasets}
  \label{tab:dataset_info}
  \centering
    \begin{tabular}{@{}lcccc@{}}
      \toprule
      Dataset & Tasks & Classes & Samples & Domain division \\
      \midrule
      NTU RGB+D\cite{nturbgd} & 40 & 60 & 56,880 & Users \\
      Toyota Smarthome\cite{toyota} & 7 & 55 & 8,848 & Scenes \\
      ETRI-Activity3D-LivingLab\cite{etri} & 40 & 60 & 6,589 & Hybird \\
      \bottomrule
    \end{tabular}
\end{table}

\subsection{Domain Replay Incremental Feature Training (DRIFT)}
\begin{figure}[tb]
  \centering
  \includegraphics[width=0.9\linewidth]{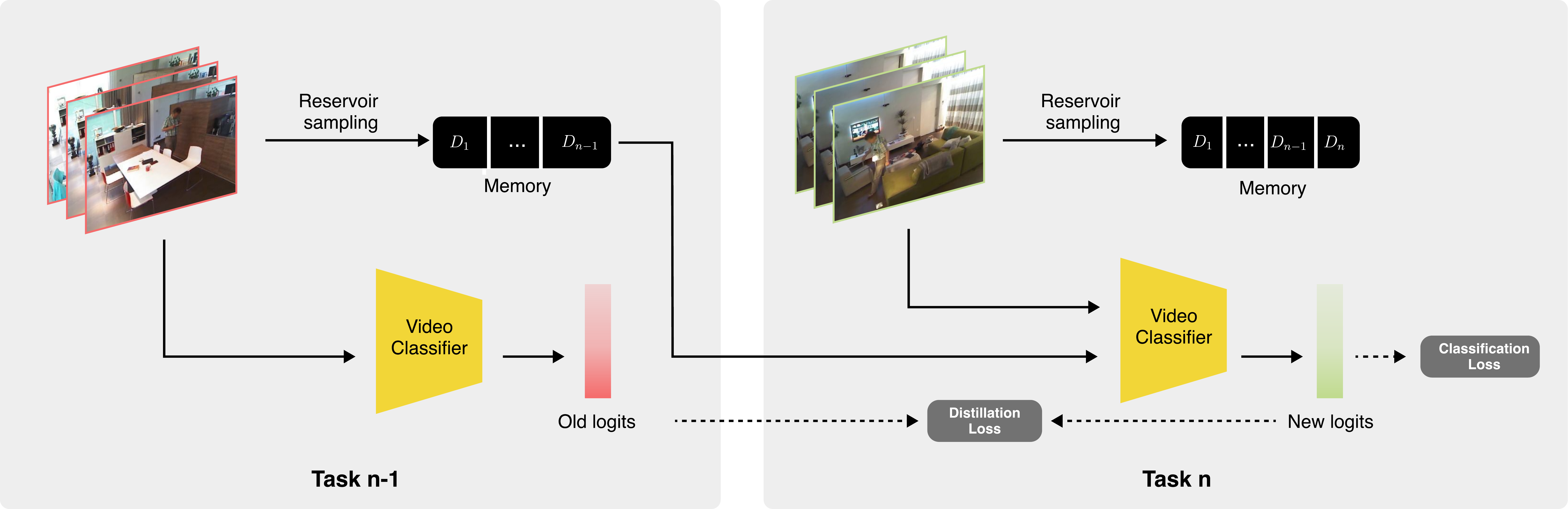}
  \caption{DRIFT leverages reservoir sampling to address the task-agnostic challenge, ensuring that each domain is equally likely to be represented in the memory. Furthermore, DRIFT incorporates a dual-loss strategy to maintain robust adaptability and memory retention across varied domains.}
  \label{fig:method}
\end{figure}

Domain Incremental Learning (DIL) in video understanding presents unique challenges compared to Class-Incremental Learning (CIL), particularly in mitigating catastrophic forgetting across diverse domains while maintaining a fixed set of action classes. In DIL, the primary challenge lies in adapting to shifts in domain characteristics that preserve class labels but significantly alter underlying feature distributions. This adaptation can inadvertently lead to the degradation of previously learned domain-specific features. \cref{fig:tsne} illustrates this challenge through t-SNE clustering, revealing that within home environments, variations in the scene domain lead to more pronounced domain shifts compared to variations between different action classes. Moreover, in real-world scenarios, especially in home environments, data often arrives in a continuous stream without explicit task boundaries, complicating timely model updates and effective management of the sample buffer necessary to counteract forgetting and ensure stable performance across evolving domains.

To address these challenges, we propose a baseline strategy named DRIFT (Domain Replay Incremental Feature Training). \cref{fig:method} delineates the proposed framework, designed for domain incremental learning in video classification. Unlike traditional rehearsal-based approaches such as iCaRL\cite{rebuffiICaRLIncrementalClassifier2017}, LwF\cite{liLearningForgetting2018}, GEM\cite{zhaoMaintainingDiscriminationFairness2020}, DRIFT leverages reservoir sampling to avoid relying on task boundaries to populate the buffer as the training progresses. It prevents data bias by ensuring the sample buffer fairly represents all tasks and domains, and it dynamically adapts to continuous data streams without relying on predefined task boundaries.

Besides, DRIFT employs a dual-loss strategy to enhance model adaptability and memory retention across varying domains.  The Classification Loss minimize the cross-entropy loss for both current domain data and previous domain retained in the memory buffer, promoting consistent performance across all domains and aligning the model's present capabilities with its cumulative learning experiences. The Classification Loss is defined as:

\begin{equation}
\mathcal{L}_{\text{class}} = - \sum_{(V_{t,j}, y_{t,j}) \in \mathcal{D}_t \cup \mathcal{B}} \sum_i y_{t,j}^i \log \left(f_{\theta}(i|V_{t,j})\right)
\end{equation}

To ensure the model's fidelity to previously acquired domain knowledge while learning new domain specifics, DRIFT employs the KD Loss. This loss component mitigates the forgetting of characteristics from older domains, crucial as the model assimilates new information. The KD Loss measures the divergence between the outputs from the previous model iteration and the current model's outputs, maintaining the stability of learned representations. The KD Loss is quantified by:

\begin{equation}
\mathcal{L}_{\text{KD}} = - \sum_{j=1}^{|\mathcal{D}_t|} \left( f_{\theta_{t-1}}(V_{t,j}) / {T} \right) \log \left( f_{\theta_t}(V_{t,j}) / {T} \right)
\end{equation}

where \( |\mathcal{D}_t| \) represents the number of samples in the current task \( t \), where \( V_{t,j} \) is the \( j \)-th video in task \( t \). The parameters \( \theta_{t-1} \) and \( \theta_t \) indicate the model's parameters before and after training on task \( t \) respectively. The function \( f_{\theta}(V_{t,j}) \) denotes the output of the model for the video \( V_{t,j} \) under these parameters. The temperature parameter \( T \), which is used to soften the output logits, enhances the distillation process by making it more effective and helps preserve the integrity of the model's learned behaviors.

Integrating these components, the overall loss function for DRIFT is:
\begin{equation}
\mathcal{L} = \mathcal{L}_{\text{class}} + \lambda \mathcal{L}_{\text{KD}}
\end{equation}
where \( \lambda \) is a hyper-parameter balancing the importance of the KD Loss in the overall training objective.


\section{Experiments and Results}

\subsection{Experimental Setups}
\subsubsection{Comparison baselines.}
We first establish performance boundaries using two baseline methods. The Naïve method serves as the theoretical lower bound, wherein all weights of the pre-trained model are freely adjusted by fine-tuning on novel tasks without any strategies to prevent catastrophic forgetting. Conversely, the Joint training method represents the theoretical upper bound, assuming the availability of all task data simultaneously. We further explore several state-of-the-art regularization-based methods—LwF\cite{liLearningForgetting2018}, EWC\cite{kirkpatrickOvercomingCatastrophicForgetting2017}, and BiC\cite{wuLargeScaleIncremental2019}—that aim to control updates to the network weights through knowledge distillation or the addition of regularization terms. For replay-based methods, we implement iCaRL\cite{rebuffiICaRLIncrementalClassifier2017} and DER\cite{buzzegaDarkExperienceGeneral2020}, which mitigate catastrophic forgetting by storing and replaying a subset of past data during training to maintain the model's performance on previously learned tasks.

\subsubsection{Implementation Details}
All experiments are conducted on RTX 4090 GPUs. To ensure fair comparisons, we employ the same classifier backbone as proposed in \cite{villaVCLIMBNovelVideo2022a}, specifically a TSN model~\cite{DBLP:conf/eccv/WangXW0LTG16} with a ResNet-50 backbone pre-trained on ImageNet, configured to utilize $N=8$ segments per video. Input images are resized to $224 \times 224$ and normalized to the range of [0, 1], consistent with the pre-training setting. Models are trained using the SGD optimizer \cite{DBLP:journals/corr/Ruder16} with a learning rate of $1 \times 10^{-3}$ and momentum of 0.9 across all experiments. We use a batch size of 16 and train for 50 epochs. All methods use a softmax temperature of $T = 2$. For method-specific configurations, EWC employs a regularization factor of $3 \times 10^{3}$ on each dataset. DER employs horizontal flips as its data augmentation. In the case of BiC, the split ratio of the validation set for bias correction is set to 0.1.

\subsection{Results and Comparisons}

\begin{figure}[tb]
  \centering
  \begin{subfigure}[b]{\textwidth}
    \centering
    \includegraphics[width=0.8\textwidth]{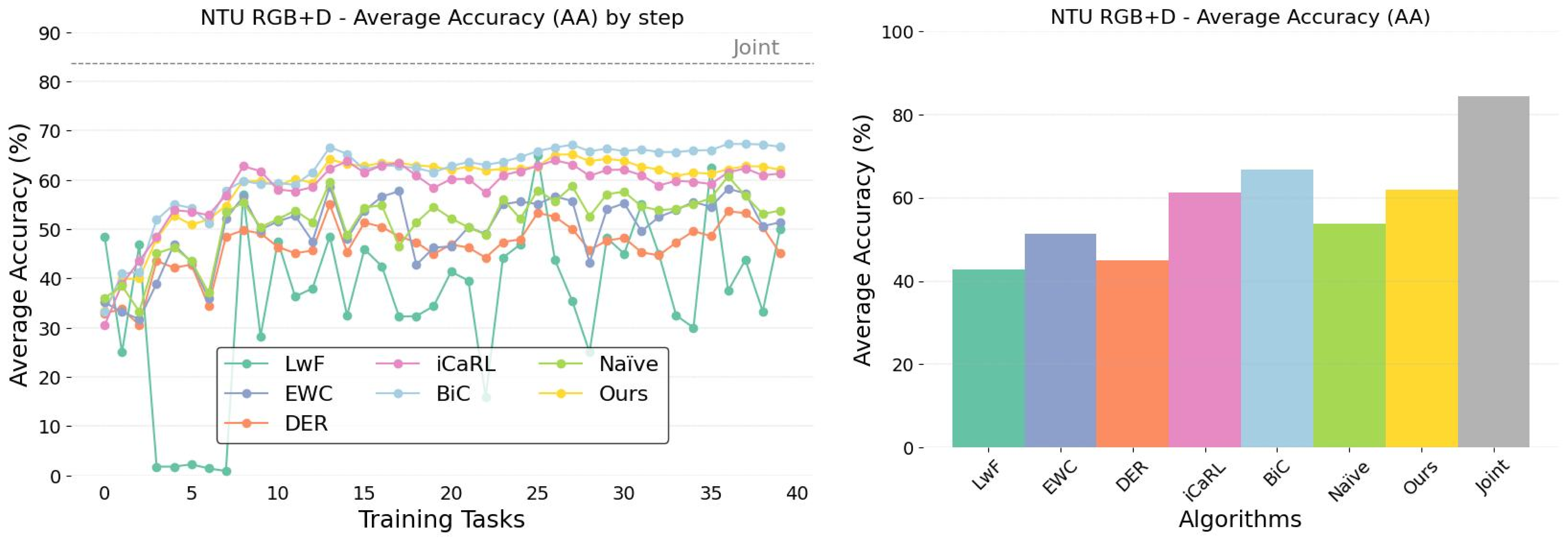}
    \caption{}
    \label{fig:chart1}
  \end{subfigure}
  
  \begin{subfigure}[b]{\textwidth}
    \centering
    \includegraphics[width=0.8\textwidth]{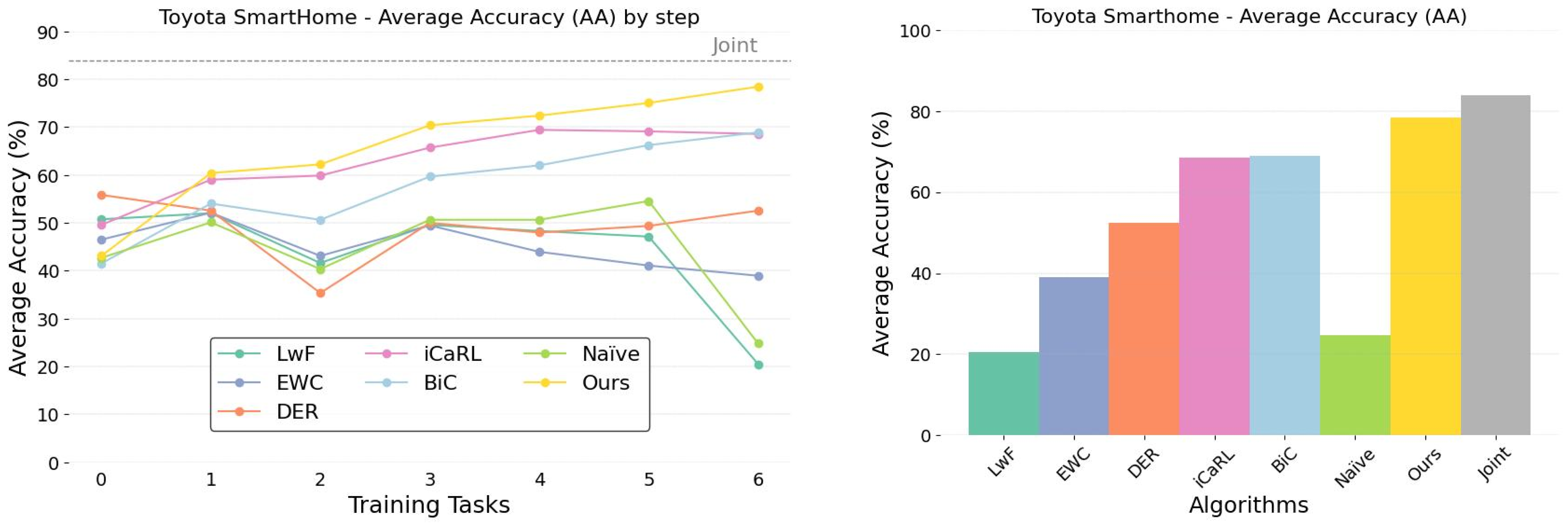}
    \caption{}
    \label{fig:chart2}
  \end{subfigure}
  
  \begin{subfigure}[b]{\textwidth}
    \centering
    \includegraphics[width=0.8\textwidth]{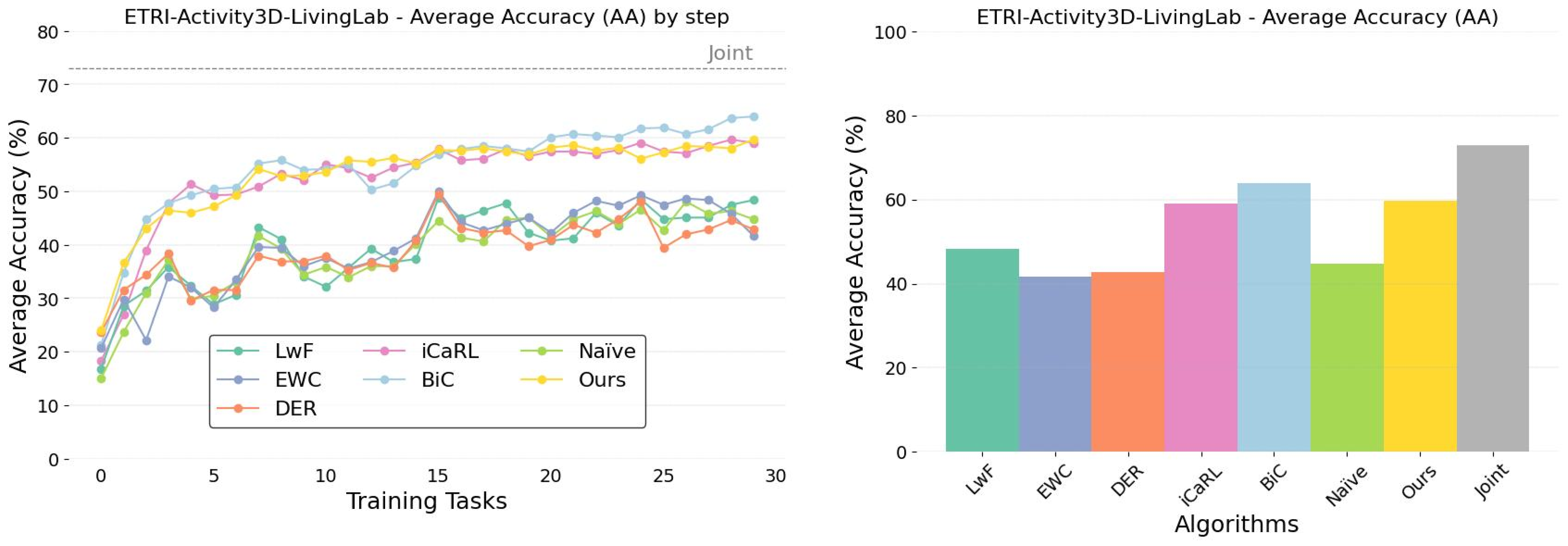}
    \caption{}
    \label{fig:chart3}
  \end{subfigure}
  
  \caption{Performance comparison of continual learning methods across three proposed benchmarks. Panels (a), (b), and (c) illustrate the performance dynamics and final average accuracies of various algorithms on the NTU RGB+D, Toyota Smarthome, and ETRI-Activity3D-LivingLab datasets, respectively.}
  
  \label{fig:combined}
\end{figure}

\begin{table}[tb]
  \caption{Performance Metrics of Continual Learning Algorithms on Diverse Benchmarks: Evaluating Average Accuracy (AA) and ackward Transfer (BWF)}
  \label{tab:overall}
  \centering
  \setlength{\tabcolsep}{6pt}
  \begin{tabular}{@{}lccc@{}}
    \toprule
    & \begin{tabular}[c]{@{}c@{}}Benchmark 1:\\ NTU RGB+D\\ (Users)\end{tabular} & \begin{tabular}[c]{@{}c@{}}Benchmark 2:\\ Toyota Smarthome\\ (Scenes)\end{tabular} & \begin{tabular}[c]{@{}c@{}}Benchmark 3:\\ ETRI-Activity3D-Livinglab\\ (Hybrid)\end{tabular} \\ 
    \midrule
    Metrics       & \textbf{AA$\uparrow$} & \textbf{AA$\uparrow$} & \textbf{AA$\uparrow$} \\ 
    \midrule
    Naïve                & 53.77 & 24.79 & 44.79 \\ 
    LwF\cite{liLearningForgetting2018}           & 50.0 & 20.49 & 48.36\\ 
    EWC\cite{kirkpatrickOvercomingCatastrophicForgetting2017}           & 51.39 & 38.96 & 41.67 \\ 
    DER\cite{buzzegaDarkExperienceGeneral2020}           & 45.09 & 52.55 & 42.86 \\ 
    iCaRL\cite{rebuffiICaRLIncrementalClassifier2017}         & 61.3 & 68.58 & 59.08 \\ 
    BiC\cite{wuLargeScaleIncremental2019}           & 66.72 & 68.9 & 63.99 \\ 
    \textbf{Ours}        & 62.05 & 78.45 & 59.67 \\ 
    \midrule
    Joint         & 84.39 & 83.9 & 73.07 \\ 
    \midrule
    \midrule
    Metrics       & \textbf{BWF$\downarrow$} & \textbf{BWF$\downarrow$} & \textbf{BWF$\downarrow$} \\ 
    \midrule
    Naïve                & -1.32 & 47.68 & -5.44 \\ 
    LwF\cite{liLearningForgetting2018}           & -9.71 & 54.73 & -5.59 \\ 
    EWC\cite{kirkpatrickOvercomingCatastrophicForgetting2017}           & -3.97 & 35.02 & -1.47 \\ 
    DER\cite{buzzegaDarkExperienceGeneral2020}           & 0.33 & 10.98 & -1.17 \\ 
    iCaRL\cite{rebuffiICaRLIncrementalClassifier2017}         & -5.60 & 11.13 & -6.68 \\ 
    BiC\cite{wuLargeScaleIncremental2019}           & -7.10 & 8.40 & -10.50 \\ 
    \textbf{Ours}        & -6.71 & 3.54 & -5.33 \\ 
    \midrule
    Joint         & - & - & - \\ 
    \bottomrule
  \end{tabular}
\end{table}

\cref{fig:combined} illustrates the performance of various continual learning methods across three benchmarks: NTU RGB+D, Toyota Smarthome, and ETRI-Activity3D-LivingLab. For each benchmark, the left chart shows the Average Accuracy (AA) trend over training tasks, while the right chart displays the final AA after completing all tasks. 

In the NTU RGB+D dataset, most methods show initial fluctuations but stabilize after a certain number of tasks, suggesting that stable temporal features can be learned from a relatively small number of domains. However, even the best continual learning strategies exhibit a significant gap (17.67\%) compared to the joint training upper bound. The Toyota Smarthome dataset demonstrates more pronounced performance differences among methods, indicating that scene-based domain shifts pose greater challenges than user-based shifts. This is evidenced by the steeper decline in the Naive method's performance compared to the NTU RGB+D dataset. The ETRI-Activity3D-LivingLab dataset shows overall lower performance across methods, which may be attributed to backbone model limitations or dataset-specific challenges such as distant subjects and subtle inter-class differences.

The comprehensive experimental metrics shown in \cref{tab:overall}. Consistent with previous findings in continual learning, replay-based methods (iCaRL, BiC, and our approach) outperform regularization-based methods (LwF, EWC). This superiority likely stems from replay methods' ability to explicitly preserve information from previous tasks, crucial in domain-incremental learning where data distributions evolve over time. Our proposed method demonstrates competitive performance across all three benchmarks, notably surpassing other methods in the Toyota Smarthome dataset with 78.45\% accuracy. This indicates our baseline's effectiveness in handling scene-based continual learning challenges. While BiC shows slightly better accuracy in NTU RGB+D and ETRI-Activity3D-Livinglab by 4.67 and 4.32 respectively, this may be attributed to BiC's bias correction mechanism being particularly effective in certain domain shift scenarios. Regarding Backward Forgetting (BWF), our method exhibits favorable performance, especially in the Toyota Smarthome dataset (3.54), suggesting effective retention of previously learned knowledge. While our method may slightly underperform compared to BiC, it notably does not require knowledge of task identifiers, offering a significant advantage in dynamically adapting to and balancing across new domains while effectively preserving previously learned knowledge

\subsection{Buffer Size Analysis}
\begin{table}[]
\caption{Memory size analysis}
\label{tab:memory-tab}
\centering
\setlength{\tabcolsep}{7pt}
\begin{tabular}{lccc}
\hline
                                           & \textbf{Memory Size} & \textbf{AA$\uparrow$}   & \textbf{BWF$\downarrow$}  \\ \hline
Joint                                      & -           & 83.9  & -    \\ 
iCaRL\cite{rebuffiICaRLIncrementalClassifier2017}                                         & 550         & 68.58 & 11.13 \\
BiC\cite{wuLargeScaleIncremental2019}                                        & 550         & 68.9  & 8.40 \\ \hline
\multicolumn{1}{l}{\multirow{3}{*}{Ours}} & 1100        & 83.17 & -2.8 \\
\multicolumn{1}{l}{}                       & 550         & 78.45 & 3.54 \\
\multicolumn{1}{l}{}                       & 225         & 74.71 & 6.11 \\ \hline
\end{tabular}%
\end{table}

In continual learning, fixed memory size leads to a decrease in stored samples per task as the number of tasks grows, weakening the model's ability to retain knowledge of previous tasks and causing catastrophic forgetting\cite{delangeContinualLearningSurvey2021}. We conducted experiments with varying memory sizes to assess model robustness under resource constraints and determine the minimum effective memory capacity, which is crucial for continual learning in the video domain due to high storage requirements.

\cref{tab:memory-tab} presents a comparative analysis of the performance of various methods under different memory constraints (1100, 550, 225). The experimental results demonstrate that the performance of all methods deteriorates as the memory size decreases, which aligns with the expectation that limited memory capacity leads to more severe forgetting\cite{rebuffiICaRLIncrementalClassifier2017}. However, our proposed method consistently outperforms the baseline methods, iCaRL and BiC, across all memory settings. Notably, even with the most stringent memory constraint of 225, our method surpasses iCaRL and BiC with a memory size of 550 in terms of both Average Accuracy (AA) and Backward Transfer (BWT) metrics. Furthermore, when utilizing a memory size of 550 (corresponding to 15.6\% of the training set), our method achieves a performance level that is comparable to or even exceeds the non-incremental learning setting (Joint), attaining an impressive AA of 83.17\%. These results underscore the efficacy of our proposed approach in alleviating catastrophic forgetting, showcasing its robustness in memory-constrained scenarios.

\subsection{Ablation Studies}

\subsubsection{Ablation Studies on Various Loss Functions}

We conduct ablation studies on loss functions for continual learning using the Toyota Smarthome dataset (\cref{tab:loss_tab}). The classification loss alone achieves 63.8\% accuracy but high forgetting (BWF 8.63), indicating strong new task learning but poor knowledge retention. Conversely, knowledge distillation loss alone significantly reduces forgetting (BWF 0.62) at the cost of lower accuracy (52.22\%). Combining $\mathcal{L}_\text{class}$ and $\mathcal{L}_\text{KD}$ yields the best performance (78.45\% accuracy, BWF 3.54), demonstrating a synergistic effect in balancing new learning and knowledge preservation. These findings demonstrate the effectiveness of dual loss approaches in our proposed baseline method, aligning with the design principles seen in established continual learning frameworks such as iCaRL and LwF.

\begin{table}[h]
\centering
\caption{Ablation Studies on Various Loss Functions Using the Toyota Smarthome Dataset}
\label{tab:loss_tab}
\setlength{\tabcolsep}{7pt} 
\begin{tabular}{cccc}
\hline
$\mathcal{L}_{\text{class}}$ & $\mathcal{L}_{\text{KD}}$ & \textbf{AA$\uparrow$} & \textbf{BWF$\downarrow$} \\ \hline
$\checkmark$ & $\times$ & 63.8 & 8.63 \\
$\times$ & $\checkmark$ & 52.22 & 0.62 \\
$\checkmark$ & $\checkmark$ & 78.45 & 3.54 \\ \hline
\end{tabular}
\end{table}

\subsubsection{Impact of Different Sampling Strategies}

We investigated the impact of different sampling strategies on incremental learning performance (\cref{tab:sampling}). We compared three strategies: (a) random selection, (b) the sample management strategy proposed by iCaRL, which selects samples closest to the class mean in the feature space for each class, and (c) entropy-based sampling, which prioritizes the selection of samples with higher entropy values, as they are considered more informative and representative of the data distribution.

All three strategies store 550 samples from old domains. The incremental learning results are shown in Table 5. As incremental learning progresses, the random selection strategy achieves the best performance in most tasks, followed by the iCaRL sample management strategy, and finally, the entropy-based sampling strategy. After the final task, the performance of the random selection strategy is 9.9\% higher than iCaRL and nearly 12.5\% higher than entropy-based sampling.

These results suggest that the random selection strategy exhibits superior performance in incremental learning. The iCaRL sample management strategy, which relies on proximity to class centers, performs slightly better than entropy-based sampling but is inferior to random selection. The relatively small performance gap between random selection and iCaRL indicates that our proposed method is not highly sensitive to the specific sample selection strategy.

\begin{table}[]
\centering
\caption{Incremental learning results on Toyota Smarthome for different sample management strategies.}
\label{tab:sampling}
\setlength{\tabcolsep}{7pt} 
\begin{tabular}{lccccccc}
\hline
& 1 & 2 & 3 & 4 & 5 & 6 & 7 \\ \hline
Entropy\cite{shewry1987maximum} & 43.10 & 52.34 & 54.78 & 62.85 & 61.04 & 66.67 & 65.92 \\
iCaRL\cite{rebuffiICaRLIncrementalClassifier2017} & \textbf{49.58} & 59.02 & 59.87 & 65.71 & 69.43 & 69.11 & 68.58 \\
Random & 43.10 & \textbf{60.40} &\textbf{ 62.21} & \textbf{70.38} & \textbf{72.40} & \textbf{75.05} &\textbf{ 78.45} \\ \hline
\end{tabular}%
\end{table}

\section{Conclusion}

In this work, we addressed the challenge of recognizing daily human actions in dynamic home environments by formalizing the Video Domain Incremental Learning (VDIL) problem. We introduced a novel benchmark with three domain split types (user, scene, hybrid) to evaluate the challenges of evolving domains in unconstrained home settings. Our proposed baseline method, leveraging replay and reservoir sampling techniques without domain labels, outperforms most existing continual learning approaches across these benchmarks. This work provides a foundation for developing robust video understanding models capable of adapting to the ever-changing nature of home environments. Future research could explore few-shot video domain incremental learning to address the practical constraints of label acquisition in real-world scenarios, further enhancing the applicability of VDIL methods in everyday home settings.

%
%
\bibliographystyle{splncs04}
\bibliography{main}
\end{document}